\documentclass[a4paper]{article}

\usepackage{INTERSPEECH2018}
\usepackage{amsthm}
\usepackage{amsmath}
\DeclareMathOperator*{\argmax}{arg\,max}
\usepackage{url}
\usepackage[utf8x]{inputenc} %IMPT: this will render spanish special characters etc.

\theoremstyle{definition}
\newtheorem{exmp}{Example}[section]

\title{Low-Latency Neural Speech Translation}
\name{Jan Niehues, Ngoc-Quan Pham, Thanh-Le Ha, Matthias Sperber and Alex Waibel}
\address{
Institute for Anthropomatics and Robotics\\
KIT - Karlsruhe Institute of Technology, Germany}
\email{firstname.lastname@kit.edu}

\begin{document}

\maketitle
\begin{abstract}
Through the development of neural machine translation, the quality of machine translation systems has been improved significantly. By exploiting advancements in deep learning, systems are now able to better approximate the complex mapping from source sentences to target sentences.
But with this ability, new challenges also arise. An example is the translation of partial sentences in low-latency speech translation. Since the model has only seen complete sentences in training, it will always try to generate a complete sentence, though the input may only be a partial sentence. 
We show that NMT systems can be adapted to scenarios where no task-specific training data is available. Furthermore, this is possible without losing performance on the original training data. We achieve this by creating artificial data and by using multi-task learning. After adaptation, we are able to reduce the number of corrections displayed during incremental output construction by 45\%, without a decrease in translation quality.
\end{abstract}

\section{Introduction}

Neural machine translation (NMT) is currently the state-of-the-art in machine translation, significantly improving translation quality in text translation \cite{WMT} as well as in speech translation \cite{IWSLT}, where the translation input is the output from a speech recognizer. The main strength of neural machine translation is improved output fluency compared to traditional approaches, such as rule-based or statistical machine translation. 

However, while the model is able to capture more complex dependencies between the source and target languages, it relies heavily on training data examples to do so. As a consequence, the model lacks the robustness at test time to handle data that is fundamentally different from what was seen in training. There are several scenarios where this can be observed. 
For example, if the input is incorrectly cased, or if a different dialect is presented at test time which has different spelling or phrasings.

In this work, we will concentrate on a speech translation use case in which the translation system is required to provide an initial translation in real time, before the complete sentence has been spoken. 
To this end, \cite{Niehues2016LT} presented an approach where partial sentences are translated and later replaced if necessary with the translations of the complete sentences. 
While we focus on this use case, the results of this work can be easily adapted to other use cases where there are differences between the training and testing scenarios.

When applying partial sentence translation to neural machine translation systems, we encounter the problem that the MT system has only been trained on complete sentences, and thus the decoder is biased to generate complete target sentences. When receiving inputs which are partial sentences, the translation outputs are not guaranteed to exactly match with the input content, which can be seen in Example \ref{motivationExample}.
We observe that the translation is often ``fantasized'' by the model to be a full sentence, as would have occurred in the training data. In the example below, although the English input ends with \textit{`all of'}, the system generates the translation \textit{`todo el mundo'} (Engl. \textit{`all of the world'}). In other cases, the decoder can fall into an over-generation state and repeat the last word several times (eg. `debería, debería, debería').

\begin{exmp}
\label{motivationExample}
Examples of challenges in using NMT to translate spoken utterances.
\begin{center}
\begin{tabular}{ll}
English:&I encourage all of \\
Spanish:&yo animo a todo el mundo . \\ \hline
English:& now , I should \\
Spanish:& ahora debería , debería , debería .
\end{tabular}
\end{center}
\end{exmp}

In this work, we aim to remedy the problem of partial sentence translation in NMT. 
Ideally, we want a model that is able to generate appropriate translations for incomplete sentences, without any compromise during other translation use cases. 
Our approach involves using multi-task learning and the automatic generation of parallel corpora in which both the source and the target sentences are incomplete sentences.

\section{Related Work}

The main topic of our work is adapting to different types of inputs for neural machine translation. Previous works have focused on domain mismatch between the training data and test data~\cite{Kobus2017}. In the case of speech translation, the model may only be exposed to specific issues arising from speech recognition outputs during test time. Since speech input can carry over errors from the ASR system to translation, it is necessary to adapt the model to noisier circumstances. To handle this scenario, previous work has proposed introducing artificially corrupted inputs at training time \cite{Sperber2017a} or direct training on lattices produced by the speech recognition system \cite{Sperber2017}.

Multi-task learning has commonly been used in various NLP problems to jointly train a single model for several well-established NLP tasks, reducing overhead and improving performance. Such implementations can be seen using the encoder-decoder model with attention mechanism~\cite{Luong2015}, in which a single model is trained for part-of-speech tagging, named-entity tagging and machine translation simultaneously~\cite{Niehues2017}.

Regarding low-latency speech translation, various approaches to translating small text segments exist using statistical phrase-based models~\cite{sridhar2013segmentation,oda2014optimizing,shavarani2015learning} or neural networks~\cite{Gu2017}. Due to the fact that the whole input sentence is not available, it is necessary to find a compromise between  translation quality and latency. The decoding process of the neural models also needs to be changed to deal with a stream of inputs, which is non-negligible. It is also possible to use the revision strategy to update the partial translations, which has been implemented in practical systems~\cite{Niehues2016LT}.

\section{Low-Latency Speech Translation}

In the practice of simultaneous speech translation, translation quality is not the only criterion; it is also important to produce a translation for a spoken utterance in real-time and at low latency. Since a speaker's utterance can be arbitrarily long, it is necessary for the translation system to start operating before the speaker stops, in which case the system input will consist of incomplete spoken segments instead of full sentences.

We explore the translation revision method of~\cite{Niehues2016LT}, which has been successfully applied to statistical translation systems. The key idea of this method is that the system iteratively revises translations by re-translating new messages sent by the speech recognition component. These newly sent messages are either replacements of or concatenations to previous ones. As a result, the user sees the translation continually updated in the interface. 

For example, for the sentence \textit{`I encourage all of you'}, the system first receives only the beginning of the sentence \textit{`I'}, with the intermediate translation being \textit{`yo'}. Afterwards, it receives an update which is the continuation of the previous one: \textit{`I encourage all of'}. The resulting translation from a typical neural model would be \textit{`yo animo a todo el mundo'}, hypothesizes a final word. Finally, the whole source sentence is available, and the MT system will update the translation of the sentence to \textit{`yo animo a todos ustedes'}.

As can be seen from the above example, in the last translation step the interface has to update the words \textit{`todo el mundo'} for \textit{`todos ustedes'}, which was generated only when the full sentence was available. As a result, we experience a delay which comes from the second to last translation step, which is longer than necessary. The interface also suffers from the update, since nearly half of the sentence needs to be replaced. Despite the fact that the final translation quality in the end does not depend on the processing of each segment, the intermediate translation outputs may change drastically due to source sequence updates. The problem is exacerbated by the fact that a neural machine translation model trained with normal parallel corpora is not able to flexibly generate translation for partial input segments, which were not available during training. The aim of our work is to build an online machine translation models which minimizes the number of words which need to be corrected until the full sentence has been seen. We aim to minimizing such criteria while maintaining translation quality for the complete utterance.

\section{Partial Translation}

As motivated in the introduction, an out-of-the-box NMT system struggles with partial input sentences. In order to improve the flexibility of the model, we investigate generating parallel corpora in which the input and output are also partial sentences. Subsequently, we adjust the training process to make use of the data in order to build a single system that is proficient at translating partial as well as full sentences.

\subsection{Generating Partial Parallel Corpora}
\label{task}

In order to build a system that is good at translating partial sentences, we need to build partial sentence training data. Since such data is not available, we investigated methods to build an artificial training set from standard parallel data. This has the advantage that the methods can be easily applied to any language pair and domain and no new data has to be collected.

Creating the source data is straightforward. Given a source sentence $S=s_1 \ldots s_I$, we can generate $I$ input samples $S^{(i)}=s_1 \ldots s_{i}$ by selecting the first $i$ words. The challenge arises from defining the correct translation for this source string. Since we are using this data in a low-latency speech translation system, the translation of the partial sentence should meet several conditions. First, it should be as long as possible in order to minimize the latency of the system. If we always used only the first word, we would not improve latency over a system that waits until the sentence is finished.
Furthermore, to minimize the number of corrections, the translation of $S^{(i)}$ should be a substring of $S^{(i')}$
 for all $i' > i$. Thus the translation of $S^{(i)}$ should be a substring of the reference translations.
 
One possible solution is to take the reference translation of the whole sentence. But, it is unrealistic to be able to generate the whole target sentence from only a single word in the source string. Therefore, we investigated two methods to select a reasonable substrings from the reference translation.

The first method is motivated by the idea that the translation should constantly generate longer target sequences when receiving longer source segments. Furthermore, word reordering may exist between two languages, for many languages sentence structure is similar. Consequently, a first approximation is to use the same proportion of words from the reference translation as we have from the source sentence.

One problem in this case is that we introduce additional noise. If the word order is different, we force the system to guess the words coming next in the source sentence. To avoid this problem, we first generate a word alignment using Giza++ \cite{Och2003} between the source and target sentences. Then, we select the longest prefix of the reference so that no target words in the prefix are aligned to source words that are not in the partial sentences:
\begin{equation}
T^{(i)}=\argmax_{j \in J}\{t_1\ldots t_j | \vee j' \le j: a(j') \le i\}
\end{equation}

\subsection{Training Process}

\paragraph*{Multi-task training} Given the artificially produced training data, a first step is to train a model on the newly created partial sentence data and use it for speech translation only. Since both tasks are very similar, we first pre-train a standard NMT system and then fine-tune the system to translate partial sentences.

The disadvantage of this approach is that the performance on complete sentences might drop, since NMT models tend to rapidly forget what they have learned before. In order to have a system that is able to generate high quality translations of both complete sentences and partial sentences, we opt to use multi-task learning, treating these as two separate tasks. In our approach, we randomly subsample the partial sentence training data to make it the same size as the original training data, so that the model can put equal emphasis on both tasks. The mixed training data then has twice as many sentences as the baseline system, but significantly less than the system using all partial sentences. Then, we fine-tune the NMT system on both tasks: translating complete sentences as well as the partial counterparts.

\paragraph*{Sequence level optimization} Beside multi-task learning, we can also guide the search operation of the model so that the generated output is better matched to the source input. We use reinforcement learning with policy gradient methods~\cite{williams1992simple,ranzato2015sequence} to train the model to maximize the GLEU score~\cite{wu2016google}, which is the combination of $n$-gram precision and recall. This reward function restricts the model from generating sentences that are too long. Since this method is known to have high variance gradients, we follow the method in~\cite{rennie2016self}, which estimates a baseline using greedy search to reduce the variance. 

\section{Experimental Results}

 \begin{table*}[!htb]
  \begin{center}
   \begin{tabular}{l|c|c|c|c|c|c|c} \hline 
 \hline 
 System & Valid (tst2011) & Test (tst2012) & \multicolumn{2}{|c|}{TEDTest Partial} & \multicolumn{3}{c}{SLT (tst2010)}\\
 & BLEU & BLEU & BLEU & length (tokens) & BLEU & Word Up & Mssg. Up. \\ \hline
Baseline & 36.86 & 31.33 & 26.66 & 509K & 25.97 & 182K & 15.0K\\
Partial & 35.45 & 30.29 & 29.48 & 375K & 25.54 & 98K & 11.8K \\ 
Multi-task & 37.05 & 31.27 & 30.09 & 376K & 26.00 & 101K & 12.0K \\ \hline 
Align. ref. & 37.13 & 31.06 & 30.29 & 371K & 26.30 & 98K & 11.5K \\ 
\hline
RL & 37.21 & 31.25 & 30.08 & 540K & 26.61 & 179K & 15.1K \\ 
RL + Multi & 37.50 & 31.21 & 30.31 & 377K & 26.77 & 82K & 11.5K \\
\hline \hline
   \end{tabular}
 \caption{\label{result:enes} Results for English to Spanish}
 
  \end{center}
 \end{table*}

We evaluate the method on three different languages pairs: English-Spanish, English-French and German-English. 

\subsection{System description}

For all experiments, we trained systems on the Europarl~\cite{Koehn2005} and the WIT-TED corpora \cite{Cettolo2012WIT} and tested on test sets from the IWSLT evaluation campaign. All systems were adapted to the TED domain by fine-tuning on the in-domain TED data. For the English$\rightarrow$Spanish and English$\rightarrow$French directions, we also optimize the models towards GLEU scores~\cite{wu2016google} using reinforcement learning (RL). 
For partial sentence translation (both data generation and training), we utilize only the TED corpus. We used the OpenNMT-py toolkit~\cite{opennmt} to train the systems. For each language pair, we jointly trained BPE \cite{Sennrich2016} for the source and target languages. 

 \begin{table}[!htb]
  \begin{center}
   \begin{tabular}{l|c|c|c|c|c} \hline 
 \hline 
 System & \multicolumn{2}{|c|}{tst2010}  & \multicolumn{3}{c}{SLT(tst2010)}\\
 & Final & Mix & BLEU & Word Up & Mssg. Up.\\ \hline 
Baseline & 34.11 & 31.18 & 23.84 & 216K & 16.3K \\ 
Multi & 34.40 & 34.71 & 23.83 & 128K & 13.5K\\ \hline
RL & 35.08 & 34.09 & 24.31 & 140K & 15.0K\\ 
RL +Multi & 34.84 & 42.51 & 24.23 & 99K & 12.1K\\ 
\hline \hline
   \end{tabular}
 \caption{\label{result:enfr} Results for English to French}
 
  \end{center}
 \end{table}

\subsection{Evaluation metrics}

We evaluated the translation quality using the BLEU score \cite{Papineni2002}. Since the ASR output uses automatic sentence segmentation, we need to re-segment the translation to fit the reference translations. Therefore, we used the method described in \cite{Matusov2005}, where the automatic translation is re-segmented in a way that minimizes the word error rate to the reference.

In addition, we also need to measure the extent to which we are able to reduce the number of corrections in the spoken language translation (SLT) system. To do so, we roll out all updates from the ASR system and translate each. 
For each updated translation, we measure the overlap between pairs of consecutive updates $s_t$ and $s_{t+1}$ and calculate the amount of re-writing necessary to produce $s_{t+1}$ after $s_t$.
Specifically, the number of corrected words is calculated by the length of the translation of $s_t$, minus the length of the common prefix both translations.
As illustrated in the example in Section 3, the final update would lead to $3$ corrected words (Word Up). Since an intermediate word change will force the user to reread all following words, our metric also counts all words following the first corrected word as corrected. We also report the number of messages where at least one word is corrected (Messg. Up.).

\subsection{Experiments}

\paragraph*{Initial results} Our initial results on the English$\rightarrow$Spanish translation task are shown in Table \ref{result:enes}. We report results in BLEU on the test and validation set with full sentence translation. Next, we report results on the test data with all possible prefixes, and finally, results on the ASR output. 
In the initial experiments, we use length ratio to determine the length of the reference for the partial translations as described in Section \ref{task}.
The baseline system is only trained on complete sentences. All other systems use the baseline system, and continue training using different strategies. The system "Partial" is fine-tuned on all partial sentences. As shown in the first two lines of Table \ref{result:enes}, the final translation quality drops significantly  by ${\sim}1$ BLEU point. 
On the other hand, the BLEU score calculated on only partial sentences improves by ${\sim}3$ BLEU points. As shown by the number of tokens, the length of translations is reduced by 25\%. So, a major problem of the baseline system is that it generates translations that are too long for the partial sentences. 
When testing on the ASR output in the last two columns, we see that the translation quality of the final hypothesis also drops, but the number of words which are updated is reduced by 45\%. Also, the number of messages where at least one word changed is reduced by 20\%. 
The system in the third line uses multi-task learning. In this case, the system is trained to perform both tasks: translation of partial and full sentences. Using this technique, we can combine the advantages of both models and maximize the translation quality of the final hypothesis, while minimizing the number of updates. The system has the same translation quality as the baseline system, with the same reduction in updates as the partial system.

\paragraph*{Performance w.r.t the artificial data} In the second set of experiments, we analyzed the use of the artificial data. We used the alignment-based method to generate the references for the partial sentences. In this case, we again fine-tuned used multi-task learning. As shown in the results in Table \ref{result:enes}, there is no clear performance difference between the two approaches. When translating text input, the system using length-ratio references is better, while the system using alignment-based methods is better on partial sentence and speech translation. Since the length-ratio based method is simpler, we used this approach for the remainder of this work.

\paragraph*{Sequence-level optimization} Finally, we also used reinforcement learning (RL) to optimize the performance of the system directly towards BLEU. These systems are first trained using cross-entropy and continue training using reinforcement learning. Here again, we have a baseline system trained only on the full sentences, and a multi-task system trained on both the full and partial sentences (final two lines of Table \ref{result:enes}). As above, we observe that with multi-task learning, we do not lose performance on full sentences, while we can significantly reduce the number of updates. In this case, the number of words updated is further reduced, reaching more than 50\% less than the baseline.

\paragraph*{English$\rightarrow$French}
We also performed experiments using two English to French systems, which are summarized in Table \ref{result:enfr}. We again have a baseline system trained with cross-entropy, and a baseline system where training is continued with RL. The models were evaluated on the full sentences and the mixed set of complete and partial sentences, as well as on the ASR output. 
Similar to the other translation directions, we improved translation performance on partial sentences and reduced the number of rewritten tokens for the SLT output by using multi-task learning. 
Interestingly, using reinforcement learning also helped us improve the performance on partial sentences. The RL criteria evaluates the n-gram precision as well as the recall of the translation, which is punished when the generated output is too long. Both methods can be combined to achieve the overall best performance, which reduced the rewritten tokens by up to 50\% without compromising translation performance.

\paragraph*{German to English}
Finally, we also performed experiments on English to German as shown in Table \ref{result:deen}. Again, we can improve the number of rewrites needed to produce the final output. For this language pair, however, the number of updates messages is only slightly reduced. The reason for this might be the larger reordering needed between the two language pairs.

 \begin{table}[!htb]
  \begin{center}
   \begin{tabular}{l|c|c|c} \hline 
 \hline 
 System & \multicolumn{3}{c}{SLT(tedX2015)}\\
 & BLEU & Words Up. & Messg. Up \\ \hline
Baseline & 15.52 & 246K & 23.6K \\
Multi-task & 15.64 & 172K & 23.1K \\
\hline \hline
   \end{tabular}
 \caption{\label{result:deen} Results for German to English}
 
  \end{center}
 \end{table}

\subsection{Examples}

In addition to the evaluation in the last section, improvements using our approach can be seen through examples for English-Spanish and German-English, shown in Table \ref{example:deen}.

In both cases, we see that the baseline system is not able to generate translations for very short sequences. In this case, the last word is repeated several times. In addition, since the NMT system is tested on input it is not accustomed to, we see that the NMT decoder relies more heavily on language modeling information and completes the sentence in a way that is typical in the target language, regardless of the source input. For example, we see the added \textit{and so on} in the first message for the German to English system. The multi-task system, however, has been trained to handle partial sentences and is therefore able to generate a correct translation. 

 \begin{table}[!htb]
  \begin{center}
   \begin{tabular}{ll} \hline 
 \multicolumn{2}{c}{English to Spanish} \\
 \hline 
Input: & now, \\
Baseline: & ahora ,\\
Multi-task: & ahora ,\\ \hline
Input : & now, I should\\ 
Baseline: & ahora debería , debería , debería .\\
Multi-task: & ahora debería \\ \hline
Input : & now, I should men\\ 
Baseline: & ahora debería hombres hombres .\\
Multi-task: & ahora debería \\ \hline
Input : & now, I should mention that this \\
Baseline: & ahora debería mencionar esto .\\
Multi-task: & ahora , debo mencionarlo .\\
\hline \hline
\multicolumn{2}{c}{German to English} \\
 \hline 
Input: & Und \\
Baseline: & And and and and and so on.\\
Multi-task: & And \\ \hline
Input : & Und ich habe\\ 
Baseline: & And I have\\
Multi-task: & And I have \\ \hline
Input : & Und ich empfehle Ihnen\\ 
Baseline: & And I recommend you to you\\
Multi-task: & And I recommend you \\ \hline
Input : & Und ich empfehle Ihnen . \\
Baseline: & And I recommend you .\\
Multi-task: & And I recommend you .\\
\hline \hline
   \end{tabular}
\caption{\label{example:deen} Examples for English-Spanish and German-English}
 
  \end{center}
 \end{table}

Finally, another interesting point is how the systems handle punctuation. The baseline model for German to English is only able to generate the correct translation if the sentences ends with a punctuation mark. This can be seen in the two last examples, which contain the same words, but only the second has punctuation. 
The multi-task system, in contrast, is able to generate the correct translation before the input is correctly punctuated. 
While most errors happen in very short (one or two words) partial sentences, longer partial sentences can also be problematic because of issues like punctuation, which suggests that ignoring short sentences is not a proper solution.

\section{Conclusion}
Low latency translation is important for real-time speech translation systems. 
To address this challenge, we improve upon a mechanism to translate partial speech input and make updates in real-time. 
Our main contribution is to propose a simple method to deal with scenarios where data at inference time is different from the training data, which can be resolved with adaptation. 
We first showed that using simple techniques to generate artificial data are effective to get more fluent output with less correction. 
We also illustrated that multi-task learning can help adapt the model to the new inference condition, without losing the original capability to translate full sentences. 
Combining these two ideas, we are able to maintain high quality translation at low latency, minimizing the number of corrected words by 45\%, which significantly improves user experience for practical applications.

\section{Acknowledgements}
This work was supported by the Carl-Zeiss-Stiftung.

\bibliographystyle{IEEEtran}

\bibliography{mybib}

\end{document}